\documentclass[11pt]{article}

\usepackage[]{acl}

\usepackage{times}
\usepackage{latexsym}
\usepackage[T1]{fontenc}
\usepackage[utf8]{inputenc}
\usepackage{microtype}
\usepackage{inconsolata}
\usepackage{graphicx}
\usepackage{booktabs}
\usepackage{tikz}
\usepackage{multirow}
\usepackage{array}
\usepackage{xcolor}
\usepackage{tikz}
\usetikzlibrary{positioning, arrows.meta}

\usepackage[most]{tcolorbox}
\definecolor{rqblue}{HTML}{000099}

\newtcolorbox{rqbox}[2][]{
  enhanced, breakable,
  colback=rqblue!5!white,
  colframe=rqblue,
  coltitle=white,
  fonttitle=\bfseries\itshape,
  title=#2,
  attach boxed title to top left={yshift=-2mm, xshift=3mm},
  boxed title style={colback=rqblue, sharp corners},
  sharp corners=south,
  left=8pt, right=8pt, top=8pt, bottom=6pt,
  #1
}

\title{LLM Safety Alignment in Low-Resource Languages: A Systematic Literature Review}

\author{
  Valdini Douglace Lemofouet\textsuperscript{1} \quad
  Blessing Ngozi Uzor\textsuperscript{1} \quad
  Paula Chikaodinaka Anyanwu\textsuperscript{1} \quad \\
  \textbf{Danielle Blanche Kapsa\textsuperscript{1}} \quad
  \textbf{Sukairaj Hafiz Imam\textsuperscript{2}} \quad 
  \textbf{P Sam Sahil\textsuperscript{7}} \quad
  \textbf{Abigail Oppong\textsuperscript{3}} \quad
  \textbf{Tassallah Abdullahi\textsuperscript{4}} \quad \\
  \textbf{Clemencia Siro\textsuperscript{5}} \quad
  \textbf{Idris Abdulmumin\textsuperscript{6}} \quad
  \textbf{Seid Muhie Yimam\textsuperscript{7}} \quad
  \textbf{Shamsuddeen Hassan Muhammad\textsuperscript{8,2}} \quad
  \\[0.5em]
  \textsuperscript{1}African Institute for Mathematical Sciences (AIMS), Cameroon \quad
  \textsuperscript{2}Bayero University Kano \quad \\
  \textsuperscript{3} Independent Researcher \quad
  \textsuperscript{4}Brown University \quad 
   \textsuperscript{5}Centrum Wiskunde \& Informatica \quad \\
  \textsuperscript{6}University of Pretoria \quad
  \textsuperscript{7}University of Hamburg \quad
  \textsuperscript{8}Imperial College London \quad
  }

\begin{document}
\maketitle

\begin{abstract}
Large Language Models (LLMs) have achieved substantial progress in safety alignment, yet their safety guarantees remain significantly weaker in low-resource and multilingual settings than in high-resource languages. In this paper, we conduct a Systematic Literature Review (SLR) of LLM safety alignment in low-resource languages by adopting the PRISMA 2020 methodology. Out of roughly 1,500 papers identified from Semantic Scholar, arXiv, and OpenAlex, 50 relevant studies have been selected and analyzed. Our review is organized around four themes: safety alignment methods, multilingual safety risks, evaluation benchmarks, and cross-lingual transferability. We further propose a taxonomy of safety alignment approaches based on three adaptation mechanisms: data adaptation, objective optimization, and mechanistic alignment. Across literature, translated English benchmarks fail to sufficiently represent culturally rooted harms, and multilingual models are more vulnerable to cross-lingual jailbreaks, code-switching attacks, and safety degradation in underrepresented languages. These failures are driven by several key factors, including uneven multilingual pre-training coverage, insufficient native-language preference data, poor transfer of safety representations, and a lack of culturally aware evaluation frameworks. The review also notes that many low-resource languages, especially African languages, have fewer safety benchmarks available than other multilingual regions. Overall, the results reveal a persistent multilingual safety gap, and suggest that future progress will require culturally grounded benchmarks, participatory data collection, balanced multilingual pre-training, and scalable multilingual alignment methods.
\end{abstract}
\section{Introduction}
\label{sec:introduction}

In recent years, Large Language Models (LLMs) have been increasingly deployed in domains such as education, healthcare, governance, and digital communication, raising concerns about their safety, reliability, and alignment with human values. 
A language model is considered safe if it avoids generating harmful, illegal, toxic, dangerous, or biased content, while remaining useful to users. Recent advances in safety alignment have introduced several techniques for reducing harmful, toxic, biased, or otherwise unsafe model behavior. These techniques include reinforcement learning from human feedback (RLHF), constitutional AI, safety fine-tuning, and adversarial red teaming~\citep{Ouyang2022RLHF,Bai2022ConstitutionalAI,perez-etal-2022-red}. 
As these systems continue to grow in capability and adoption, ensuring their safe and responsible behavior has become a major research priority. 

Most existing alignment techniques and evaluation frameworks have been designed primarily for high-resource languages, particularly English \cite{yong-etal-2025-state}. These approaches often depend on large-scale annotated datasets, benchmarks, and an extensive human feedback pipeline, resources that remain limited or unavailable for many underrepresented languages.

Safety alignment learned in English does not transfer reliably to other languages. Recent studies show that LLMs trained to reject harmful instructions in English may comply with equivalent prompts written in low-resource languages~\cite{yong2023lowresource}.  Previous studies also suggest that these failures are highly related to the imbalanced coverage of multilingual pre-training~\cite{shen2024languagebarrier}. In practice, these kinds of gaps can translate into weaker safety protections for speakers of underrepresented languages. For example, \citet{inuwa2025openai} presented examples of an LLM deployed in Hausa producing false and unsafe recommendations for toxic local products that were presented as safe for humans.

In order to better understand this emerging area, this paper presents a Systematic Literature Review (SLR) of safety alignment in low-resource languages using the PRISMA 2020 framework.  We synthesize existing research across four dimensions: safety alignment methods, multilingual safety risks, evaluation benchmarks and cross-lingual transferability. We further propose a taxonomy of safety alignment approaches by their main adaptation mechanisms and identify key challenges limiting effective safety transfer across languages. In this review, we highlight current research trends, methodological limitations, and future directions for building safer and more inclusive multilingual language models. 
This review is guided by the following research questions:

\begin{itemize}
  \item \textbf{RQ1:} How are LLM safety alignment methods adapted for low-resource languages?
  \item \textbf{RQ2:} What safety risks, adversarial behaviors, and cultural harms emerge in multilingual LLM settings?
  \item \textbf{RQ3:} What datasets, benchmarks, and evaluation frameworks exist for safety alignment in low-resource languages?
  \item \textbf{RQ4:} What factors affect cross-lingual transfer of safety alignment to low-resource languages?
\end{itemize}

\begin{figure*}[h!]
    \includegraphics[width=0.9\textwidth]{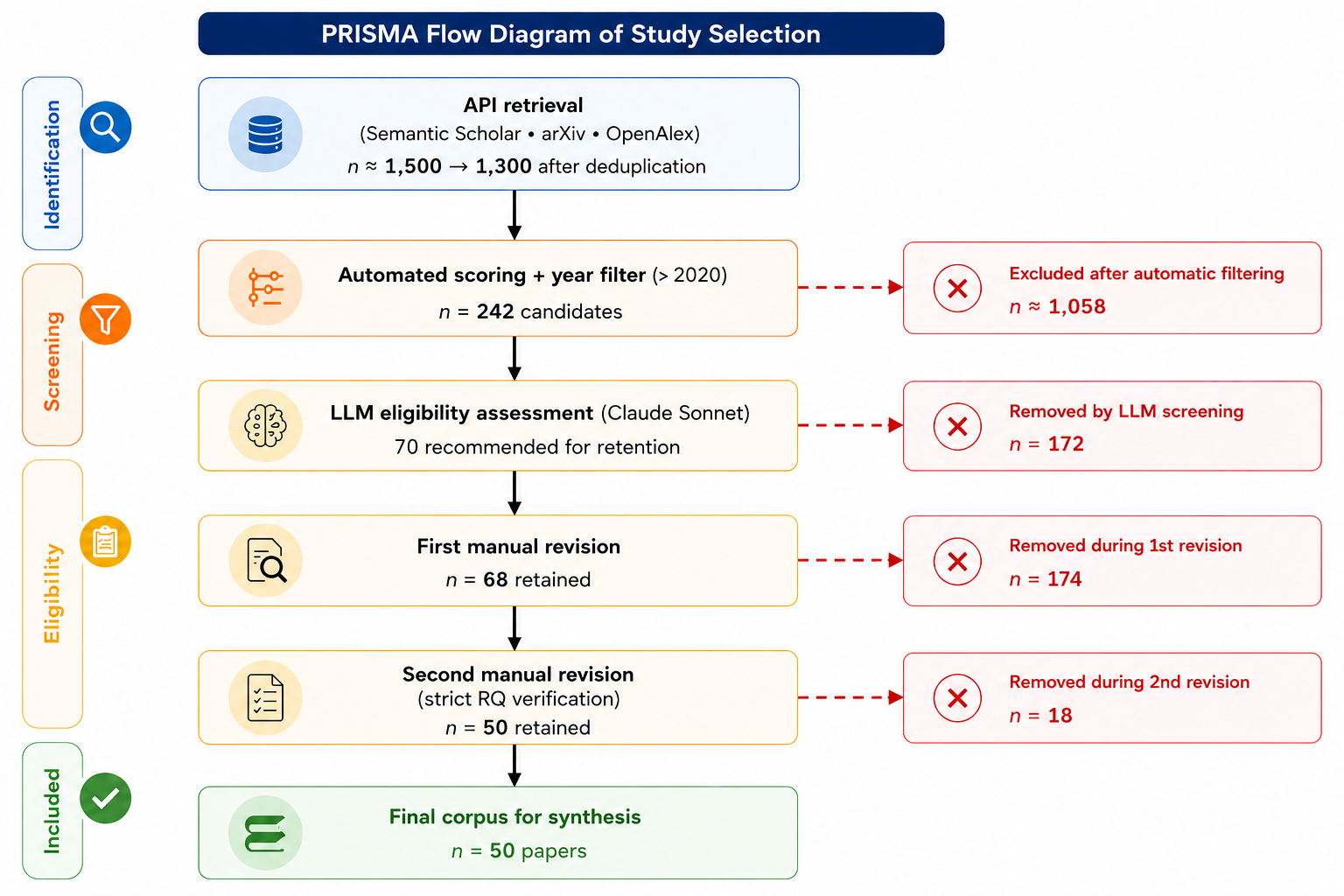}
    \caption{\textbf{PRISMA flow diagram}. 1{,}500 records were narrowed through deduplication, automated scoring, LLM-based eligibility screening, and two rounds of manual revision producing 50 papers for synthesis.}
    \label{fig:prisma}
\end{figure*}

\section{Related Work}

Several studies have been conducted on the safety of multilingual LLMs. Some of these works aimed to analyze and review the field. Among these, \citet{yong2025statemultilingualllmsafety} conducted a descriptive systematic review of nearly 300 publications, revealing a persistent and growing language gap in LLM safety research. Also, \citet{banerjee2026bridgingmultilingualsafetydivide} synthesized findings for Global South languages: "safety guardrails weaken sharply on low-resource and code-mixed inputs". They also proposed parameter‑efficient steering and participatory workflows that enable communities to define and mitigate harm. \citet{shen2024languagebarrier} examines the safety challenges of LLMs in multilingual settings. They observed that LLMs tend to generate unsafe responses much more often when a malicious prompt is written in a lower-resource language. 

These works share four limitations: (1) none follows a formal SLR methodology (explicit research questions, inclusion/exclusion criteria, PRISMA flow), (2) African languages are subsumed under generic “low‑resource” categories, (3) findings are not structured into actionable dimensions (methods, risks, benchmarks), and (4) they do not synthesise mechanistic explanations for cross‑lingual transfer failures.

Our work addresses these gaps by presenting the first systematic literature review (PRISMA) dedicated exclusively to LLM safety alignment in African and low‑resource languages.

\section{Methodology}
\label{sec:methodology}

We performed a systematic literature review based on the PRISMA 2020 guidelines~\cite{Page2021PRISMA}. The PRISMA flow diagram is shown in Fig. \ref{fig:prisma}. Relevant articles were collected from Semantic Scholar, arXiv, and OpenAlex using searches involving queries belonging to the following three keyword sets: (i) \emph{technology} (LLM, instruction tuning, foundation models), (ii) \emph{safety} (alignment, jailbreak, adversarial attack, toxicity, RLHF, DPO), and (iii) \emph{linguistic scope} (low-resource languages, multilingual, African languages, cross-lingual, code-switching, Hausa, Swahili, Yoruba, Amharic). Detailed search strings are available in Appendix~\ref{app:search}.This process generated approximately 1,500 papers, which were pared down to 1,300 after de-duplication. A Python-based filter system was used with criteria such as year cut-offs ($>$2020) and multi-faceted keyword relevancy, generating 242 potential papers to review. The inclusion criteria included the focus on safety or adversarial robustness for LLMs in multilingual and low-resource contexts, while purely English and non-empirical works were excluded. We then applied a two-step screening process. The first step involved LLM-based evaluation (Claude Sonnet) of each potential paper based on research questions and inclusion/exclusion criteria: 70 papers were selected, and after verification, none of the papers excluded by Claude were useful. We retained 50 papers in the final review set.




\section{Safety Alignment Methods for Low-Resource Languages}
\label{sec:rq1}

\begin{rqbox}{RQ1 — Adaptation of safety methods}
How are LLM safety alignment methods adapted for low-resource languages?
\end{rqbox}


\label{sec:rq1}

\begin{figure*}[h]
    \centering
    \includegraphics[width=1\textwidth]{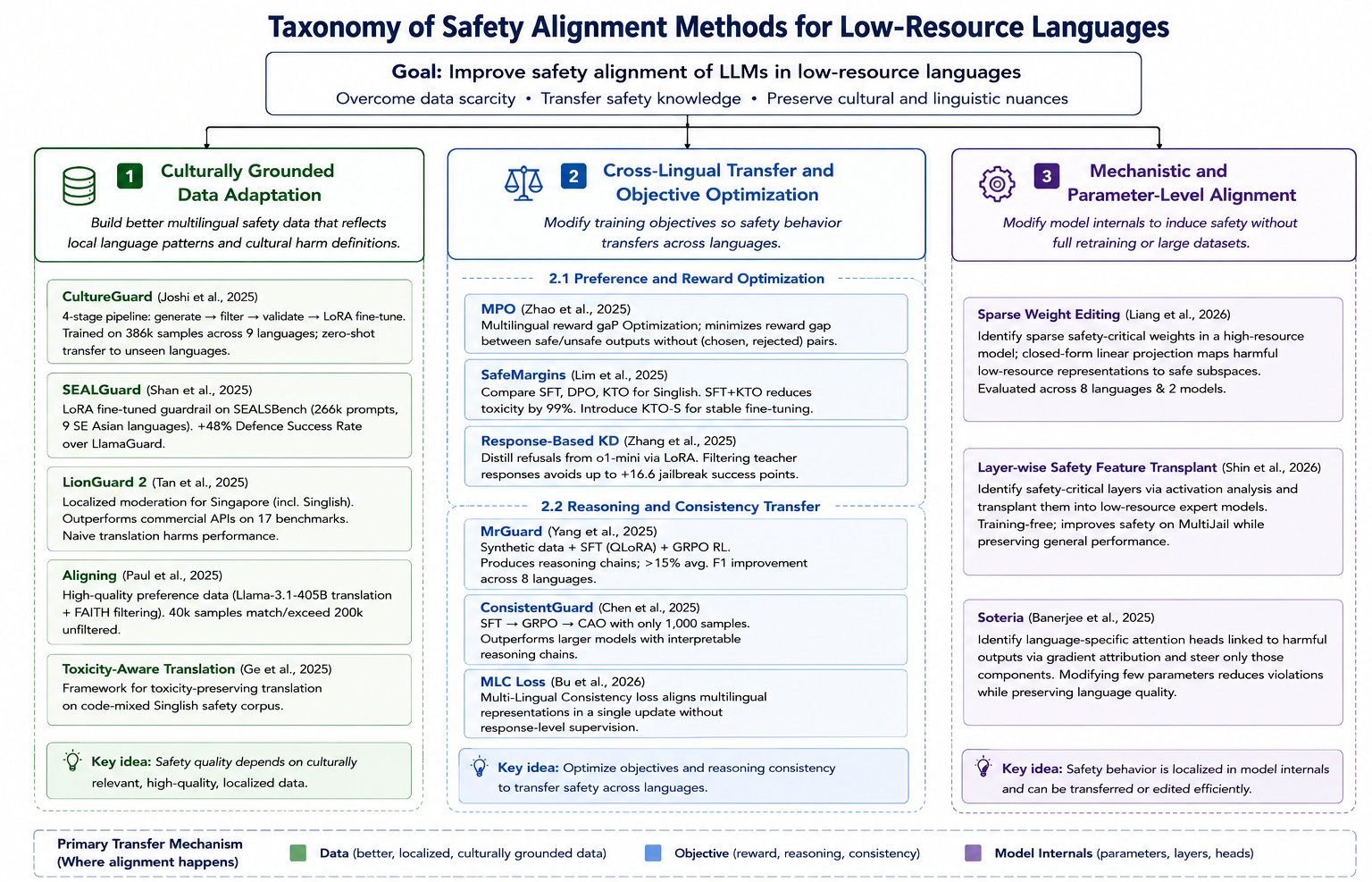}
    \caption{\textbf{Taxonomy of safety alignment methods for low-resource languages}, grouped by primary transfer mechanism: data adaptation, objective optimization, and mechanistic alignment.}    
    \label{fig:rq1}
\end{figure*}

Safety alignment for low-resource languages has become increasingly important as large language models are deployed across multilingual settings. Most safety alignment methods were designed for English and rely on large annotated datasets that do not exist for the majority of the world's languages. Applying them to low-resource settings requires adapting one of four components: \textbf{(i)} the data adaptation, \textbf{(ii)} the optimization objective, \textbf{(iii)} the model's internal parameters, or \textbf{(iv)} data.


\subsection{Culturally Grounded Data Adaptation}

The most direct response to low-resource safety failures is to build better multilingual training data. Methods in this family share a common premise: safety quality is primarily a function of how well the training data covers local language patterns and cultural harm definitions.

\citet{Joshi2025CultureGuardTC} addressed this issue through \textbf{CultureGuard}, a four-stage pipeline that generates a culturally grounded safety dataset in target languages using LLMs, filters it with a safety classifier, validates it with native speakers, and fine-tunes a guard model with LoRA. Trained on 386k samples across nine languages, including Hindi and Thai, the resulting model outperforms models trained on translated English data on non-English benchmarks and achieves zero-shot transfer to unseen languages.

\citet{Shan2025SEALGuardST} introduce \textbf{SEALGuard}, a multilingual guardrail designed to improve the safety alignment across for Southeast Asian languages. \textbf{SEALGuard} is fine-tuned(LoRA) on SEALSBench, (266k-prompt benchmark of locally sourced and translated harmful prompts across nine languages). The model improves Defence Success Rate by 48\% over  LlamaGuard, confirming that English-centric guardrails fail on region-specific jailbreak attempts. Another angle is taken by \citet{Tan2025LionGuard2B} with \textbf{LionGuard~2},a lightweight, data-efficient, localized multilingual content moderation classifier for Singapore's linguistic context. By incorporating Singapore-specific harm categories and code-mixed Singlish data, the model outperforms commercial moderation APIs across 17 benchmarks. The authors also find that naively translated training data reduces performance, reinforcing the case for localized data construction.

\citet{paul2025aligning} demonstrate that data quality matters more than scale for alignment(Hindi). They translate preference data using Llama-3.1-405B, which preserves code, formulas, and URLs that standard machine translation corrupts, then filter outputs with FAITH-based quality metrics before running a two-stage alignment process consisting of Supervised Fine-Tuning (SFT) and Direct Preference Optimization (DPO). Forty thousand filtered samples match or exceed the alignment performance of 200k unfiltered ones. In parallel, \citet{Bell2025TranslateTD} investigate translation-based multilingual toxicity detection pipelines, where the initial step is to translate potentially offensive content to English before classifying it. The authors demonstrate that translate-and-classify frameworks always perform better than multilingual classifiers when dealing with most low-resource languages.

Apart from multilingual moderation datasets, other researchers have also developed resources for foundational safety alignment datasets in low-resource languages.\citet{Huang2025TFDAC} present the \textbf{Tibetan Foundation Dataset (TFD)}, which is a massive dataset in the Tibetan language that can be used in all stages of LLM training processes, from pre-training to safety alignment and reasoning. 

For safety alignment purposes, some methods for translating existing datasets suffer from a problem of preserving intent (harmful, toxic). It is in this context that \citet{Ge2025ToxicityAwareFP} proposes a framework for toxicity-preserving translation, demonstrated on a code-mixed Singlish safety corpus.

\subsection{Cross-Lingual Transfer and Objective Optimization}

When multilingual safety data is scarce, a second strategy is to modify the training objective so that safety behaviour learned in one language transfers to others. This family splits into two complementary directions: reward-based optimization and reasoning-based consistency.

\subsubsection{Preference and Reward Optimization}

\citet{Zhao2025MPOMS} address the paired preference data bottleneck with \textbf{MPO} (Multilingual reward gaP Optimization). Rather than requiring (chosen, rejected) pairs in each target language, MPO minimizes the reward gap between safe and unsafe outputs directly across languages. It consistently outperforms RLHF and DPO baselines on cross-lingual safety benchmarks under noisy conditions.

\citet{lim2025safemargins} compare SFT, DPO, and Kahneman-Tversky Optimization (KTO) for Singlish, where preference pairs are difficult to collect. Their combined SFT+KTO pipeline reduces toxicity by 99\% and introduces KTO-S, an improved regularization strategy that stabilizes fine-tuning under data scarcity. \citet{Zhang2025ResponseBasedKD} use knowledge distillation to
transfer refusal behaviour from OpenAI o1-mini into open-source multilingual models via LoRA. Jailbreak resistance improves in both high and low-resource languages, but the authors find a critical caveat: unfiltered distillation increases jailbreak success rates by up to 16.6 points because ambiguous ``boundary'' refusals from the teacher confuse the student. Filtering these responses mitigates the degradation.

\subsubsection{Reasoning and Consistency Transfer}

\citet{Yang2025MrGuardAM} introduced \textbf{MrGuard}, which combines synthetic multilingual data generation, SFT( with QLoRA), and GRPO-based reinforcement learning. A distinctive feature is that the model produces explicit reasoning chains rather than binary labels, which improves generalization across eight languages(including LRL) by more than 15\% in average F1 and maintains robustness against code-switching attacks.

 \citet{Chen2025UnlockingLS} presented \textbf{ConsistentGuard}, a guardrail model trained using only 1,000 samples. They combines supervised fine-tuning (to provide the model with initial task-specific knowledge), followed by GRPO (to promote reasoning diversity and length), and finally a CAO (constrained Alignment Optimization) to align the model’s reasoning process across different languages. The model obtained outperforms larger classifiers while providing interpretable reasoning chains, suggesting that reasoning supervision is an efficient substitute for scale.

\citet{bu2026align} proposes a resource-efficient method for improving multilingual safety alignment. Their Multi-Lingual Consistency (MLC) loss enforces directional consistency between multilingual representation vectors in a single training update, enabling simultaneous safety alignment across languages without response-level supervision in target languages. Further evaluation across languages and tasks indicates improved cross-lingual generalization, suggesting the approach as a practical solution for multilingual consistency alignment under limited supervision.

\subsection{Mechanistic and Parameter-Level Alignment}

A third line of research investigates whether safety alignment can be achieved through direct modification of model internals, thereby bypassing data collection and full retraining entirely. The hypothesis is that safety behaviour is concentrated in identifiable subsets of parameters.

In this line of research , \citet{Liang2026MultilingualSA} propose the Sparse Weight Editing (SWE), based on the finding that safety capabilities are concentrated in a sparse subset of model weights. They identify this subset in a safety-aligned high-resource model and compute a closed-form linear transformation to project harmful low-resource representations into the corresponding safety subspaces without gradient computation. SWE was evaluated across eight languages and two model families (Llama-3, Qwen-2.5). It reduces attack success rates in low-resource languages, with negligible impact on general reasoning performance.

\citet{Shin2026LayerwiseSF} takes a layer-level approach: Activation analysis identifies which transformer layers carry safety-critical features in a safety-aligned high-resource model. Those layers are then transplanted into a low-resource expert model. The resulting training-free method achieves safety gains on MultiJail while preserving performance on general benchmarks (MMMLU, BELEBELE, MGSM).

\citet{Banerjee2025SoteriaLF} localize the intervention further
with \textbf{Soteria}, which identifies language-specific attention heads responsible for harmful outputs via gradient-based attribution and steers only those components. Modifying a small fraction of parameters drastically reduces policy violations while preserving language quality, even in low-resource settings.


\section{Safety Risks in Multilingual and Low resource Language Contexts}
\label{sec:rq2}


\begin{rqbox}{RQ2 - Risks and cultural harms}
What safety risks, adversarial behaviors, and cultural harms emerge in multilingual LLM settings?
\end{rqbox}

 Although much research is focused on reducing the safety gap between HRLs and LRLs, understanding the safety risks faced by low-resource languages is crucial. Here, We present these LLM vulnerabilities in LRL settings.
 
\subsection{Cross-Lingual Jailbreaks}

A consistent body of work shows that safety vulnerabilities increase significantly in low-resource and underrepresented languages.

Early evidence from \citet{yong2023lowresource} demonstrates that simply translating harmful English prompts into low-resource languages can achieve a 79\% jailbreak success rate on GPT-4, while high-resource languages remain below 15\%. This disparity suggests that safety alignment does not generalize evenly across linguistic space.

Building on this, \citet{Deng2023MultilingualJC} identifies multilingual jailbreak challenges within LLMs and examines two potential risk scenarios: unintentional and intentional. They found that LRLs have about 3 times the likelihood of encountering harmful content as HRLs. \citet{shen2024languagebarrier} also supports these findings, noting that in low-resource languages, models not only become less safe but also follow instructions less reliably as linguistic resources decrease.

It is important to note that these weaknesses are not limited to translation attacks. In fact, \citet{Chrabaszcz2025EvaluatingLR} show that for Polish, a small proxy model can generate transferable adversarial perturbations at low cost, producing attack success rates significantly higher than those observed in English. \citet{Pattnayak2026IndicJRAJ} reach a related finding across 12 Indic languages through IndicJR: contract-bound prompts in JSON format inflate refusal counts without actually preventing jailbreaks, and safety alignment degrades consistently as language resource level decreases. \citet{Atil2025DoMT} extend this picture to ten languages, finding that both logical-expression-based and adversarial-prompt-based jailbreak methods generalize poorly across languages, with low-resource languages remaining more vulnerable.

\subsection{Fine-Tuning on New Languages as an Attack Vector}

Beyond prompting-based attacks, a more subtle risk arises from model adaptation itself.

\citet{upadhyay2025tonguetied} show that fine-tuning aligned models on new or synthetic languages, even using only benign data, can still degrade safety alignment. This suggests that learning new linguistic mappings can interfere with previously established safety constraints. For African language adaptation pipelines, this implies that introducing new linguistic domains without safety recalibration may reduce robustness.

\subsection{Code-Switching and Multilingual Blending}

Multilingual interaction further expands the attack surface through code-switching. \citet{Song2024MultilingualBL} show that mixing languages within a single prompt significantly reduces safety robustness, with bypass rates reaching 67.23\% on GPT-3.5 and 40.34\% on GPT-4. These effects are not uniform and depend on linguistic distance and prompt structure. This is particularly relevant in African contexts, where code-switching is a common communicative norm rather than an adversarial construct.

\subsection{Culturally Specific Harm}
A growing body of work also highlights that safety failures are not only linguistic but cultural in nature. For example, \citet{inuwa2025openai} shows that GPT-OSS-20B produces unsafe or misleading outputs in Hausa, including toxic product recommendations and culturally inappropriate content that can amplify hate speech. \citet{Shukla2026LostIT} further demonstrates that composite harms often change meaning during translation, leading multilingual safety systems to miss culturally contextualized unsafe content.

Beyond explicit toxicity \cite{Saeed2025SurfacingSS}, their multilingual evaluations also reveal subtle stereotype propagation and culturally dependent bias patterns (mostly in low-resource settings) that are often missed by conventional safety benchmarks.


\section{Safety Evaluation Benchmarks}
\label{sec:rq3}

\begin{rqbox}{RQ3 - Datasets and evaluation}
What datasets, benchmarks, and evaluation frameworks exist for safety alignment in low-resource languages?
\end{rqbox}

\subsection{Global Multilingual Safety Benchmarks}

Recent multilingual safety benchmarks have progressively moved beyond translation toward more culturally diverse evaluation settings. Early efforts such as \textbf{XSafety} \cite{Wang2024AllLM} introduced a benchmark of 28,000 annotated instances across 14 safety issues and 10 languages spanning multiple language families(including Hindi). Building on this direction, \textbf{LinguaSafe} \cite{Ning2025LinguaSafeAC} combines translated, trans-created, and natively written prompts across 12 languages(including Malay and Bengali). On his side  \cite{Kumar2025PolyGuardAM} introduce \textbf{POLYGUARDPROMPTS} a multilingual benchmark for the evaluation of safety guardrails, Created by combining naturally occurring multilingual human-LLM interactions and human-verified machine translations of an English-only safety dataset.

Other benchmarks focus more specifically on toxicity and adversarial robustness. \cite{Jain2024PolygloToxicityPromptsME} introduce \textbf{PolygloToxicityPrompts} a large-scale multilingual toxicity evaluation benchmark of 425K naturally occurring prompts spanning 17 languages. whereas \textbf{ML-Bench} \cite{zhao2026mlbenchguardpolicygroundedmultilingualsafety} introduces policy-grounded multilingual safety evaluation covering 14 languages, based on regional regulations.

\subsection{Regional and Culturally Localized Benchmarks}

In other side, a growing body of work focuses on culturally localized safety evaluation rather than direct translation from English benchmarks. In Southeast Asia, \textbf{SEALSBench} \cite{Shan2025SEALGuardST} and \textbf{SEA-SafeguardBench} \cite{Tasawong2025SEASafeguardBenchEA} introduce multilingual safety datasets built around regional socio-cultural risks and native-language prompts. 

Several benchmarks also target multilingual adversarial behavior and online toxicity. \textbf{SGToxicGuard} \cite{Hu2025ToxicityRB} evaluates conversational toxicity and red-teaming scenarios across Singapore's multilingual setting, while \textbf{Qorgau} \cite{Goloburda2025QorgauEL} was designed for safety evaluation in Kazakh(a low-resource language) and Russian. In South Asia, \textbf{IndicSafe} \cite{pattnayak2026indicsafebenchmarkevaluatingmultilingual} and \textbf{IndicJR} \cite{Pattnayak2026IndicJRAJ} extend evaluation to culturally grounded harms and jailbreak robustness across Indic languages. Complementing these efforts, \textbf{SEAHateCheck} \cite{Ng2026SEAHateCheckFT} and \textbf{IndoSafety} \cite{Azmi2025IndoSafetyCG} introduce hate speech and safety evaluation datasets for Southeast Asian languages and regional language varieties.

\subsection{African Language Safety Benchmarks}
Compared with other multilingual regions, benchmark resources for African languages remain limited. \textbf{LSR} \cite{fFaruna2026LSRLS} evaluates the cross-lingual refusal degradation in West African languages such as Yoruba, Hausa, Igbo, and Igala, while \cite{abdullahi2026ubuntuguard} introduces \textbf{UbuntuGuard}, the first African policy-based safety benchmark. Beyond toxicity evaluation, \textbf{Uhura} \cite{Bayes2024UhuraAB} studies truthfulness and safety constraints in African and low-resource settings.

Overall, current benchmark development shows a gradual shift toward native-language evaluation, culturally grounded harms, and multilingual adversarial testing. However, African languages remain substantially under-represented compared with other multilingual benchmark ecosystems.


\section{Cross-Lingual Transferability of Safety Alignment}
\label{sec:rq4}

\begin{rqbox}{RQ4 — Cross-lingual transfer factors}
What factors affect cross-lingual transfer of safety alignment to low-resource languages?
\end{rqbox}

A recurring finding across multilingual safety research is that safety alignment is not language-agnostic: alignment behaviors learned in high-resource languages often degrade when transferred to low-resource languages. Most studies attribute this limitation to uneven multilingual representations acquired during pre-training, where safety-relevant knowledge remains concentrated in high-resource languages.

Several prior works identify pre-training coverage as the primary bottleneck behind transfer degradation. \citet{shen2024languagebarrier} show that alignment methods  mainly improve safety in languages that are well represented during pre-training, while gains remain limited in low-resource settings. Similarly, \citet{Verma2025TheHS} show that safety-relevant features cluster around high-resource regions of the latent space, resulting in weaker safety enforcement in underrepresented languages. \citet{upadhyay2025tonguetied} further show that introducing new languages during fine-tuning can disrupt previously learned safety behavior, suggesting that multilingual adaptation itself may destabilize alignment.

Transfer is also affected by linguistic and structural differences across languages. Morphologically rich languages and underrepresented scripts often suffer from fragmented tokenization and weaker semantic representations, reducing the reliability of safety reasoning and refusal behavior. These issues become particularly visible in African and other low-resource languages, where both pre-training data and alignment supervision remain scarce.

More recent work explores whether safety transfer can be improved directly at the representation level. \citet{Liang2026MultilingualSA} propose Sparse Weight Editing (SWE), which transfers safety representations through lightweight parameter transformations, while \citet{Shin2026LayerwiseSF} improves multilingual robustness by identifying and replacing safety-critical transformer layers without full retraining. Mechanistic studies such as \citet{zhang2026transferssafetyidentifyingtargeting} further suggest that cross-lingual safety behavior may depend on a sparse set of shared multilingual safety neurons that can be selectively targeted during alignment.\citet{wang2026refusaldirectionuniversalsafetyaligned} find a shared directional structure in refusal behavior across safety-aligned languages, implying that multilingual safety alignment could depend on partially universal latent safety representations.

Recent approaches also investigate how safety transfer can scale efficiently across many languages. \citet{Bansal2025CRESTUS} show that alignment learned from a carefully selected subset of languages can generalize broadly through multilingual representation clustering. Complementing this direction, \citet{bu2026align} introduces a multilingual consistency objective that explicitly enforces alignment agreement across languages, improving transfer without requiring extensive target-language supervision.


\section{Discussion}
\label{sec:discussion}

The common denominator for these four research questions is that the prevailing methodologies used for developing safety alignment are mostly based on English-language data and have not been properly extended to accommodate the other languages of the world. This imbalance spans from the data used in pre-training to alignment procedures themselves as well as evaluation techniques.

Methodologically speaking(RQ1 and RQ4), techniques like parameter-efficient fine-tuning, data synthesis, and cross-lingual transfer might be helpful, yet they are poorly validated in the context of the African languages. As long as the pre-trained models lack coverage of these languages, the gains brought by such techniques are rather marginal. Partly because, at the pre-training stage, these models already under-represent the languages, which makes recovery of any kind of multilingual representation impossible.

A similar discrepancy appears in the safety risk landscape (RQ2). Issues such as cross-lingual jailbreak transfer, code-switching based vulnerabilities, and unintended safety degradation through benign fine-tuning into new languages have been reported, but seldom evaluated. Furthermore, culturally specific risks, such as those reported in Hausa \cite{inuwa2025openai}, are almost completely neglected in benchmarks designed with English-centric assumptions. According to \citet{Vajjala2025ThePW}, however, such limitations do not apply exclusively to models; annotation discrepancies and culturally biased benchmarks contribute to the inadequacy of evaluation measures.

The evaluation landscape (RQ3) reveals a similar pattern. While benchmarks for multiple languages continue to emerge, there is still limited representation of African languages. Translation is still the primary strategy used in the creation of datasets, although this methodology has repeatedly demonstrated its ability to manipulate meanings and mislead safety annotations.

Combined, these limitations create a reinforcing cycle whereby inadequate pre-training coverage translates into poor representation, which then limits the capabilities of the alignment methods, whereas poor benchmarks limit the ability to assess the deficiencies. To overcome this problem, we need to see progress on all fronts simultaneously, from moving away from translation-based dataset building, to better native language benchmark creation, to creating alignment methods that take this problem into consideration.

\section{Conclusion}
\label{sec:conclusion}

This systematic literature review synthesized findings from 50 studies on LLM safety alignment in low-resource language settings. The review showed that current safety alignment approaches remain largely designed for high-resource languages and do not generalize effectively across diverse linguistic and cultural contexts. Existing benchmarks rely heavily on translated datasets that often fail to capture culturally grounded harms. The findings further indicated that multilingual adaptation and fine-tuning may reduce safety performance in low-resource settings. However, this review is limited to English-language studies from major academic databases and may not fully represent locally published or recent research. Additionally, grouping languages into broad regional categories may obscure important linguistic differences. Future research should focus on native-language safety benchmarks, balanced multilingual pre-training, culturally aware evaluation methods, and participatory frameworks that actively involve African language communities in defining AI safety standards and practices.

\bibliography{custom}

@article{Page2021PRISMA,
  author = {Page, Matthew J and Moher, David and Bossuyt, Patrick M and Boutron, Isabelle and Hoffmann, Tammy C and Mulrow, Cynthia D and Shamseer, Larissa and Tetzlaff, Jennifer M and Akl, Elie A and Brennan, Sue E and Chou, Roger and Glanville, Julie and Grimshaw, Jeremy M and Hr{\'o}bjartsson, Asbj{\o}rn and Lalu, Manoj M and Li, Tianjing and Loder, Elizabeth W and Mayo-Wilson, Evan and McDonald, Steve and McGuinness, Luke A and Stewart, Lesley A and Thomas, James and Tricco, Andrea C and Welch, Vivian A and Whiting, Penny and McKenzie, Joanne E},
	title = {PRISMA 2020 explanation and elaboration: updated guidance and exemplars for reporting systematic reviews},
	volume = {372},
	elocation-id = {n160},
	year = {2021},
	doi = {10.1136/bmj.n160},
	publisher = {BMJ Publishing Group Ltd},
	URL = {https://www.bmj.com/content/372/bmj.n160},
	eprint = {https://www.bmj.com/content/372/bmj.n160.full.pdf},
	journal = {BMJ}
}

@inproceedings{
yong2023lowresource,
title={Low-Resource Languages Jailbreak {GPT}-4},
author={Zheng Xin Yong and Cristina Menghini and Stephen Bach},
booktitle={Socially Responsible Language Modelling Research},
year={2023},
url={https://openreview.net/forum?id=pn83r8V2sv},
address="New Orleans, United States"
}

@inproceedings{shen2024languagebarrier,
  title = "The Language Barrier: Dissecting Safety Challenges of {LLM}s in Multilingual Contexts",
    author = "Shen, Lingfeng  and
      Tan, Weiting  and
      Chen, Sihao  and
      Chen, Yunmo  and
      Zhang, Jingyu  and
      Xu, Haoran  and
      Zheng, Boyuan  and
      Koehn, Philipp  and
      Khashabi, Daniel",
    editor = "Ku, Lun-Wei  and
      Martins, Andre  and
      Srikumar, Vivek",
    booktitle = "Findings of the Association for Computational Linguistics: ACL 2024",
    month = aug,
    year = "2024",
    address = "Bangkok, Thailand",
    publisher = "Association for Computational Linguistics",
    url = "https://aclanthology.org/2024.findings-acl.156/",
    doi = "10.18653/v1/2024.findings-acl.156",
    pages = "2668--2680"
}

@inproceedings{upadhyay2025tonguetied,
  title = "Tongue-Tied: Breaking {LLM}s Safety Through New Language Learning",
    author = "Upadhayay, Bibek  and
      Behzadan, Vahid",
    editor = "Winata, Genta Indra  and
      Kar, Sudipta  and
      Zhukova, Marina  and
      Solorio, Thamar  and
      Ai, Xi  and
      Hamed, Injy  and
      Ihsani, Mahardika Krisna Krisna  and
      Wijaya, Derry Tanti  and
      Kuwanto, Garry",
    booktitle = "Proceedings of the 7th Workshop on Computational Approaches to Linguistic Code-Switching",
    month = may,
    year = "2025",
    address = "Albuquerque, New Mexico, USA",
    publisher = "Association for Computational Linguistics",
    url = "https://aclanthology.org/2025.calcs-1.5/",
    doi = "10.18653/v1/2025.calcs-1.5",
    pages = "32--47",
    ISBN = "979-8-89176-053-0"
}

@misc{inuwa2025openai,
  title={OpenAI's GPT-OSS-20B Model and Safety Alignment Issues in a Low-Resource Language}, 
      author={Isa Inuwa-Dutse},
      year={2025},
      eprint={2510.01266},
      archivePrefix={arXiv},
      primaryClass={cs.CL},
      url={https://arxiv.org/abs/2510.01266}, 
}

@INPROCEEDINGS{Shan2025SEALGuardST,
author = { Shan, Wenliang and Fu, Michael and Yang, Rui and Tantithamthavorn, Chakkrit },
booktitle = { 2025 2nd IEEE/ACM International Conference on AI-powered Software (AIware) },
title = {{ SEALGuard: Safeguarding the Multilingual Conversations in Southeast Asian Languages for AI-Powered Software }},
year = {2025},
volume = {},
ISSN = {},
pages = {197-206},
doi = {10.1109/AIware69974.2025.00029},
url = {https://doi.ieeecomputersociety.org/10.1109/AIware69974.2025.00029},
publisher = {IEEE Computer Society},
address = {Los Alamitos, CA, USA},
month =Nov}

@inproceedings{Tan2025LionGuard2B,
    title = "{L}ion{G}uard 2: Building Lightweight, Data-Efficient {\&} Localised Multilingual Content Moderators",
    author = "Tan, Leanne  and
      Chua, Gabriel  and
      Ge, Ziyu  and
      Lee, Roy Ka-Wei",
    editor = {Habernal, Ivan  and
      Schulam, Peter  and
      Tiedemann, J{\"o}rg},
    booktitle = "Proceedings of the 2025 Conference on Empirical Methods in Natural Language Processing: System Demonstrations",
    month = nov,
    year = "2025",
    address = "Suzhou, China",
    publisher = "Association for Computational Linguistics",
    url = "https://aclanthology.org/2025.emnlp-demos.20/",
    doi = "10.18653/v1/2025.emnlp-demos.20",
    pages = "264--285",
    ISBN = "979-8-89176-334-0"
}

@inproceedings{
Zhang2025ResponseBasedKD,
title={Response-Based Knowledge Distillation for Multilingual Jailbreak Prevention Unwittingly Compromises Safety},
author={Max Zhang and Derek Liu and Kai Zhang and Joshua Franco and Haihao Liu and Kevin Zhu},
booktitle={4th Deployable AI Workshop},
year={2026},
url={https://openreview.net/forum?id=LR6rXdJAUX},
address="Singapore Expo, Singapore"
}

@inproceedings{
lim2025safemargins,
title={Safe at the Margins: A General Approach to Safety Alignment in Low-Resource English Languages {\textendash} A Singlish Case Study},
author={Isaac Lim and Shaun Khoo and Watson Wei Khong Chua and Jessica Foo and Jia Yi Goh and Roy Ka-Wei Lee},
booktitle={Second Workshop on Language Models for Underserved Communities (LM4UC)},
year={2025},
url={https://openreview.net/forum?id=x9KuyCwksR}
}

@inproceedings{paul2025aligning,
    title = "Aligning Large Language Models to Low-Resource Languages through {LLM}-Based Selective Translation: A Systematic Study",
    author = "Paul, Rakesh  and
      Kamath, Anusha  and
      Singla, Kanishk  and
      Joshi, Raviraj  and
      Vaidya, Utkarsh  and
      Chauhan, Sanjay Singh  and
      Wartikar, Niranjan",
    editor = "Bhattacharya, Arnab  and
      Goyal, Pawan  and
      Ghosh, Saptarshi  and
      Ghosh, Kripabandhu",
    booktitle = "Proceedings of the 1st Workshop on Benchmarks, Harmonization, Annotation, and Standardization for Human-Centric AI in Indian Languages (BHASHA 2025)",
    month = dec,
    year = "2025",
    address = "Mumbai, India",
    publisher = "Association for Computational Linguistics",
    url = "https://aclanthology.org/2025.bhasha-1.6/",
    doi = "10.18653/v1/2025.bhasha-1.6",
    pages = "69--82",
    ISBN = "979-8-89176-313-5"
}

@inproceedings{Chen2025UnlockingLS,
    title = "Unlocking {LLM} Safeguards for Low-Resource Languages via Reasoning and Alignment with Minimal Training Data",
    author = "Chen, Zhuowei  and
      Zhang, Bowei  and
      Lin, Nankai  and
      Hou, Tian  and
      Wang, Lianxi",
    editor = "Adelani, David Ifeoluwa  and
      Arnett, Catherine  and
      Ataman, Duygu  and
      Chang, Tyler A.  and
      Gonen, Hila  and
      Raja, Rahul  and
      Schmidt, Fabian  and
      Stap, David  and
      Wang, Jiayi",
    booktitle = "Proceedings of the 5th Workshop on Multilingual Representation Learning (MRL 2025)",
    month = nov,
    year = "2025",
    address = "Suzhuo, China",
    publisher = "Association for Computational Linguistics",
    url = "https://aclanthology.org/2025.mrl-main.7/",
    doi = "10.18653/v1/2025.mrl-main.7",
    pages = "96--105",
    ISBN = "979-8-89176-345-6"
}

@inproceedings{Joshi2025CultureGuardTC,
    title = "{C}ulture{G}uard: Towards Culturally-Aware Dataset and Guard Model for Multilingual Safety Applications",
    author = "Joshi, Raviraj Bhuminand  and
      Paul, Rakesh  and
      Singla, Kanishk  and
      Kamath, Anusha  and
      Evans, Michael  and
      Luna, Katherine  and
      Ghosh, Shaona  and
      Vaidya, Utkarsh  and
      Long, Eileen Margaret Peters  and
      Chauhan, Sanjay Singh  and
      Wartikar, Niranjan",
    editor = "Inui, Kentaro  and
      Sakti, Sakriani  and
      Wang, Haofen  and
      Wong, Derek F.  and
      Bhattacharyya, Pushpak  and
      Banerjee, Biplab  and
      Ekbal, Asif  and
      Chakraborty, Tanmoy  and
      Singh, Dhirendra Pratap",
    booktitle = "Proceedings of the 14th International Joint Conference on Natural Language Processing and the 4th Conference of the Asia-Pacific Chapter of the Association for Computational Linguistics",
    month = dec,
    year = "2025",
    address = "Mumbai, India",
    publisher = "The Asian Federation of Natural Language Processing and The Association for Computational Linguistics",
    url = "https://aclanthology.org/2025.ijcnlp-long.144/",
    doi = "10.18653/v1/2025.ijcnlp-long.144",
    pages = "2666--2685",
    ISBN = "979-8-89176-298-5"
}

@inproceedings{bu2026align,
title={Align Once, Benefit Multilingually: Enforcing Multilingual Consistency for {LLM} Safety Alignment},
author={Yuyan Bu and Xiaohao Liu and ZhaoXing Ren and Yaodong Yang and Juntao Dai},
booktitle={The Fourteenth International Conference on Learning Representations},
year={2026},
url={https://openreview.net/forum?id=ueknOG1wXL},
address="Rio de Janeiro, Brazil"
}

@misc{abdullahi2026ubuntuguard,
   title={UbuntuGuard: A Culturally-Grounded Policy Benchmark for Equitable AI Safety in African Languages}, 
      author={Tassallah Abdullahi and Macton Mgonzo and Mardiyyah Oduwole and Paul Okewunmi and Abraham Owodunni and Ritambhara Singh and Carsten Eickhoff},
      year={2026},
      eprint={2601.12696},
      archivePrefix={arXiv},
      primaryClass={cs.CL},
      url={https://arxiv.org/abs/2601.12696}, 
}

@inproceedings{Yang2025MrGuardAM,
  title = "{M}r{G}uard: A Multilingual Reasoning Guardrail for Universal {LLM} Safety",
    author = "Yang, Yahan  and
      Dan, Soham  and
      Li, Shuo  and
      Roth, Dan  and
      Lee, Insup",
    editor = "Christodoulopoulos, Christos  and
      Chakraborty, Tanmoy  and
      Rose, Carolyn  and
      Peng, Violet",
    booktitle = "Proceedings of the 2025 Conference on Empirical Methods in Natural Language Processing",
    month = nov,
    year = "2025",
    address = "Suzhou, China",
    publisher = "Association for Computational Linguistics",
    url = "https://aclanthology.org/2025.emnlp-main.1392/",
    doi = "10.18653/v1/2025.emnlp-main.1392",
    pages = "27377--27396",
    ISBN = "979-8-89176-332-6"
}

@inproceedings{Zhao2025MPOMS,
  title = "{MPO}: Multilingual Safety Alignment via Reward Gap Optimization",
    author = "Zhao, Weixiang  and
      Hu, Yulin  and
      Deng, Yang  and
      Wu, Tongtong  and
      Zhang, Wenxuan  and
      Guo, Jiahe  and
      Zhang, An  and
      Zhao, Yanyan  and
      Qin, Bing  and
      Chua, Tat-Seng  and
      Liu, Ting",
    editor = "Che, Wanxiang  and
      Nabende, Joyce  and
      Shutova, Ekaterina  and
      Pilehvar, Mohammad Taher",
    booktitle = "Proceedings of the 63rd Annual Meeting of the Association for Computational Linguistics (Volume 1: Long Papers)",
    month = jul,
    year = "2025",
    address = "Vienna, Austria",
    publisher = "Association for Computational Linguistics",
    url = "https://aclanthology.org/2025.acl-long.1149/",
    doi = "10.18653/v1/2025.acl-long.1149",
    pages = "23564--23587",
    ISBN = "979-8-89176-251-0"
}

@inproceedings{
Deng2023MultilingualJC,
title={Multilingual Jailbreak Challenges in Large Language Models},
author={Yue Deng and Wenxuan Zhang and Sinno Jialin Pan and Lidong Bing},
booktitle={The Twelfth International Conference on Learning Representations},
year={2024},
url={https://openreview.net/forum?id=vESNKdEMGp},
address="Vienna, Austria"
}

@inproceedings{Banerjee2025SoteriaLF,
 title = "Soteria: Language-Specific Functional Parameter Steering for Multilingual Safety Alignment",
    author = "Banerjee, Somnath  and
      Layek, Sayan  and
      Chatterjee, Pratyush  and
      Mukherjee, Animesh  and
      Hazra, Rima",
    editor = "Christodoulopoulos, Christos  and
      Chakraborty, Tanmoy  and
      Rose, Carolyn  and
      Peng, Violet",
    booktitle = "Findings of the Association for Computational Linguistics: EMNLP 2025",
    month = nov,
    year = "2025",
    address = "Suzhou, China",
    publisher = "Association for Computational Linguistics",
    url = "https://aclanthology.org/2025.findings-emnlp.497/",
    doi = "10.18653/v1/2025.findings-emnlp.497",
    pages = "9347--9364",
    ISBN = "979-8-89176-335-7"
}

@inproceedings{Pattnayak2026IndicJRAJ,
    title = "{I}ndic{JR}: A Judge-Free Benchmark of Jailbreak Robustness in {S}outh {A}sian Languages",
    author = "Pattnayak, Priyaranjan  and
      Chowdhuri, Sanchari",
    editor = {Matusevych, Yevgen  and
      Eryi{\u{g}}it, G{\"u}l{\c{s}}en  and
      Aletras, Nikolaos},
    booktitle = "Proceedings of the 19th Conference of the {E}uropean Chapter of the {A}ssociation for {C}omputational {L}inguistics (Volume 5: Industry Track)",
    month = mar,
    year = "2026",
    address = "Rabat, Morocco",
    publisher = "Association for Computational Linguistics",
    url = "https://aclanthology.org/2026.eacl-industry.50/",
    doi = "10.18653/v1/2026.eacl-industry.50",
    pages = "649--668",
    ISBN = "979-8-89176-384-5"
}

@misc{Saeed2025SurfacingSS,
      title={Surfacing Subtle Stereotypes: A Multilingual, Debate-Oriented Evaluation of Modern LLMs}, 
      author={Muhammed Saeed and Muhammad Abdul-mageed and Shady Shehata},
      year={2026},
      eprint={2511.01187},
      archivePrefix={arXiv},
      primaryClass={cs.CL},
      url={https://arxiv.org/abs/2511.01187}, 
}

@inproceedings{Song2024MultilingualBL,
    title = "Multilingual Blending: Large Language Model Safety Alignment Evaluation with Language Mixture",
    author = "Song, Jiayang  and
      Huang, Yuheng  and
      Zhou, Zhehua  and
      Ma, Lei",
    editor = "Chiruzzo, Luis  and
      Ritter, Alan  and
      Wang, Lu",
    booktitle = "Findings of the Association for Computational Linguistics: NAACL 2025",
    month = apr,
    year = "2025",
    address = "Albuquerque, New Mexico",
    publisher = "Association for Computational Linguistics",
    url = "https://aclanthology.org/2025.findings-naacl.191/",
    doi = "10.18653/v1/2025.findings-naacl.191",
    pages = "3433--3449",
    ISBN = "979-8-89176-195-7"
}

@misc{Atil2025DoMT,
      title={Do Methods to Jailbreak and Defend LLMs Generalize Across Languages?}, 
      author={Berk Atil and Rebecca J. Passonneau and Fred Morstatter},
      year={2025},
      eprint={2511.00689},
      archivePrefix={arXiv},
      primaryClass={cs.CL},
      url={https://arxiv.org/abs/2511.00689}, 
}

@misc{Chrabaszcz2025EvaluatingLR,
      title={Evaluating LLMs Robustness in Less Resourced Languages with Proxy Models}, 
      author={Maciej Chrabąszcz and Katarzyna Lorenc and Karolina Seweryn},
      year={2025},
      eprint={2506.07645},
      archivePrefix={arXiv},
      primaryClass={cs.CL},
      url={https://arxiv.org/abs/2506.07645}, 
}

@inproceedings{Wang2024AllLM,
  title = "All Languages Matter: On the Multilingual Safety of {LLM}s",
    author = "Wang, Wenxuan  and
      Tu, Zhaopeng  and
      Chen, Chang  and
      Yuan, Youliang  and
      Huang, Jen-tse  and
      Jiao, Wenxiang  and
      Lyu, Michael",
    editor = "Ku, Lun-Wei  and
      Martins, Andre  and
      Srikumar, Vivek",
    booktitle = "Findings of the Association for Computational Linguistics: ACL 2024",
    month = aug,
    year = "2024",
    address = "Bangkok, Thailand",
    publisher = "Association for Computational Linguistics",
    url = "https://aclanthology.org/2024.findings-acl.349/",
    doi = "10.18653/v1/2024.findings-acl.349",
    pages = "5865--5877"
}

@misc{Ning2025LinguaSafeAC,
      title={LinguaSafe: A Comprehensive Multilingual Safety Benchmark for Large Language Models}, 
      author={Zhiyuan Ning and Tianle Gu and Jiaxin Song and Shixin Hong and Lingyu Li and Huacan Liu and Jie Li and Yixu Wang and Meng Lingyu and Yan Teng and Yingchun Wang},
      year={2025},
      eprint={2508.12733},
      archivePrefix={arXiv},
      primaryClass={cs.CL},
      url={https://arxiv.org/abs/2508.12733}, 
}

@inproceedings{
Kumar2025PolyGuardAM,
title={PolyGuard: A Multilingual Safety Moderation Tool for 17 Languages},
author={Priyanshu Kumar and Devansh Jain and Akhila Yerukola and Liwei Jiang and Himanshu Beniwal and Thomas Hartvigsen and Maarten Sap},
booktitle={Second Conference on Language Modeling},
year={2025},
url={https://openreview.net/forum?id=wbAWKXNeQ4},
address="Montreal, Canada"
}

@misc{fFaruna2026LSRLS,
      title={LSR: Linguistic Safety Robustness Benchmark for Low-Resource West African Languages}, 
      author={Godwin Abuh Faruna},
      year={2026},
      eprint={2603.19273},
      archivePrefix={arXiv},
      primaryClass={cs.CL},
      url={https://arxiv.org/abs/2603.19273}, 
}

@misc{Shukla2026LostIT,
      title={Lost in Translation? A Comparative Study on the Cross-Lingual Transfer of Composite Harms}, 
      author={Vaibhav Shukla and Hardik Sharma and Adith N Reganti and Soham Wasmatkar and Bagesh Kumar and Vrijendra Singh},
      year={2026},
      eprint={2602.07963},
      archivePrefix={arXiv},
      primaryClass={cs.CL},
      url={https://arxiv.org/abs/2602.07963}, 
}

@misc{
Tasawong2025SEASafeguardBenchEA,
title={{SEA}-SafeguardBench: Evaluating {AI} Safety in {SEA} Languages and Cultures},
author={Panuthep Tasawong and Jian Gang Ngui and Alham Fikri Aji and Trevor Cohn and Peerat Limkonchotiwat},
year={2025},
address="Rio de Janeiro, Brazil",
url={https://openreview.net/forum?id=ScVl9QlLpD}
}

@misc{Ge2025ToxicityAwareFP,
      title={Toxicity-Aware Few-Shot Prompting for Low-Resource Singlish Translation}, 
      author={Ziyu Ge and Gabriel Chua and Leanne Tan and Roy Ka-Wei Lee},
      year={2025},
      eprint={2507.11966},
      archivePrefix={arXiv},
      primaryClass={cs.CL},
      url={https://arxiv.org/abs/2507.11966}, 
}

@misc{Huang2025TFDAC,
      title={TFD: A Comprehensive Structured Tibetan Foundation Dataset for Low-Resource Language Processing and Large-Scale Modeling}, 
      author={Cheng Huang and Fan Gao and Nyima Tashi and Yutong Liu and Xiangxiang Wang and Thupten Tsering and Ban Ma-bao and Xiao Feng and Renzeg Duojie and Gadeng Luosang and Rinchen Dongrub and Dorje Tashi and Hao Wang and Yongbin Yu},
      year={2026},
      eprint={2503.18288},
      archivePrefix={arXiv},
      primaryClass={cs.CL},
      url={https://arxiv.org/abs/2503.18288}, 
}

@misc{pattnayak2026indicsafebenchmarkevaluatingmultilingual,
      title={IndicSafe: A Benchmark for Evaluating Multilingual LLM Safety in South Asia}, 
      author={Priyaranjan Pattnayak and Sanchari Chowdhuri},
      year={2026},
      eprint={2603.17915},
      archivePrefix={arXiv},
      primaryClass={cs.CL},
      url={https://arxiv.org/abs/2603.17915}, 
}

@misc{Bayes2024UhuraAB,
      title={Uhura: A Benchmark for Evaluating Scientific Question Answering and Truthfulness in Low-Resource African Languages}, 
      author={Edward Bayes and Israel Abebe Azime and Jesujoba O. Alabi and Jonas Kgomo and Tyna Eloundou and Elizabeth Proehl and Kai Chen and Imaan Khadir and Naome A. Etori and Shamsuddeen Hassan Muhammad and Choice Mpanza and Igneciah Pocia Thete and Dietrich Klakow and David Ifeoluwa Adelani},
      year={2024},
      eprint={2412.00948},
      archivePrefix={arXiv},
      primaryClass={cs.CL},
      url={https://arxiv.org/abs/2412.00948}, 
}

@inproceedings{Hu2025ToxicityRB,
  title = "Toxicity Red-Teaming: Benchmarking {LLM} Safety in {S}ingapore{'}s Low-Resource Languages",
    author = "Hu, Yujia  and
      Hee, Ming Shan  and
      Nakov, Preslav  and
      Lee, Roy Ka-Wei",
    editor = "Christodoulopoulos, Christos  and
      Chakraborty, Tanmoy  and
      Rose, Carolyn  and
      Peng, Violet",
    booktitle = "Proceedings of the 2025 Conference on Empirical Methods in Natural Language Processing",
    month = nov,
    year = "2025",
    address = "Suzhou, China",
    publisher = "Association for Computational Linguistics",
    url = "https://aclanthology.org/2025.emnlp-main.612/",
    doi = "10.18653/v1/2025.emnlp-main.612",
    pages = "12183--12201",
    ISBN = "979-8-89176-332-6"
}

@inproceedings{Goloburda2025QorgauEL,
    title = "Qor{\'{g}}au: Evaluating Safety in {K}azakh-{R}ussian Bilingual Contexts",
    author = "Goloburda, Maiya  and
      Laiyk, Nurkhan  and
      Turmakhan, Diana  and
      Wang, Yuxia  and
      Togmanov, Mukhammed  and
      Mansurov, Jonibek  and
      Sametov, Askhat  and
      Mukhituly, Nurdaulet  and
      Wang, Minghan  and
      Orel, Daniil  and
      Mujahid, Zain Muhammad  and
      Koto, Fajri  and
      Baldwin, Timothy  and
      Nakov, Preslav",
    editor = "Che, Wanxiang  and
      Nabende, Joyce  and
      Shutova, Ekaterina  and
      Pilehvar, Mohammad Taher",
    booktitle = "Findings of the Association for Computational Linguistics: ACL 2025",
    month = jul,
    year = "2025",
    address = "Vienna, Austria",
    publisher = "Association for Computational Linguistics",
    url = "https://aclanthology.org/2025.findings-acl.507/",
    doi = "10.18653/v1/2025.findings-acl.507",
    pages = "9765--9784",
    ISBN = "979-8-89176-256-5"
}

@inproceedings{Azmi2025IndoSafetyCG,
    title = "{I}ndo{S}afety: Culturally Grounded Safety for {LLM}s in {I}ndonesian Languages",
    author = "Azmi, Muhammad Falensi  and
      Al Kautsar, Muhammad Dehan  and
      Wicaksono, Alfan Farizki  and
      Koto, Fajri",
    editor = "Christodoulopoulos, Christos  and
      Chakraborty, Tanmoy  and
      Rose, Carolyn  and
      Peng, Violet",
    booktitle = "Proceedings of the 2025 Conference on Empirical Methods in Natural Language Processing",
    month = nov,
    year = "2025",
    address = "Suzhou, China",
    publisher = "Association for Computational Linguistics",
    url = "https://aclanthology.org/2025.emnlp-main.465/",
    doi = "10.18653/v1/2025.emnlp-main.465",
    pages = "9135--9166",
    ISBN = "979-8-89176-332-6"
}

@misc{Ng2026SEAHateCheckFT,
      title={SEAHateCheck: Functional Tests for Detecting Hate Speech in Low-Resource Languages of Southeast Asia}, 
      author={Ri Chi Ng and Aditi Kumaresan and Yujia Hu and Roy Ka-Wei Lee},
      year={2026},
      eprint={2603.16070},
      archivePrefix={arXiv},
      primaryClass={cs.CL},
      url={https://arxiv.org/abs/2603.16070}, 
}

@misc{Jain2024PolygloToxicityPromptsME,
      title={PolygloToxicityPrompts: Multilingual Evaluation of Neural Toxic Degeneration in Large Language Models}, 
      author={Devansh Jain and Priyanshu Kumar and Samuel Gehman and Xuhui Zhou and Thomas Hartvigsen and Maarten Sap},
      year={2024},
      eprint={2405.09373},
      archivePrefix={arXiv},
      primaryClass={cs.CL},
      url={https://arxiv.org/abs/2405.09373}, 
}

@misc{Vajjala2025ThePW,
      title={The Problem with Safety Classification is not just the Models}, 
      author={Sowmya Vajjala},
      year={2025},
      eprint={2507.21782},
      archivePrefix={arXiv},
      primaryClass={cs.CL},
      url={https://arxiv.org/abs/2507.21782}, 
}

@inproceedings{yong2025statemultilingualllmsafety,
    title = "The State of Multilingual {LLM} Safety Research: From Measuring The Language Gap To Mitigating It",
    author = "Yong, Zheng Xin  and
      Ermis, Beyza  and
      Fadaee, Marzieh  and
      Bach, Stephen  and
      Kreutzer, Julia",
    editor = "Christodoulopoulos, Christos  and
      Chakraborty, Tanmoy  and
      Rose, Carolyn  and
      Peng, Violet",
    booktitle = "Proceedings of the 2025 Conference on Empirical Methods in Natural Language Processing",
    month = nov,
    year = "2025",
    address = "Suzhou, China",
    publisher = "Association for Computational Linguistics",
    url = "https://aclanthology.org/2025.emnlp-main.800/",
    doi = "10.18653/v1/2025.emnlp-main.800",
    pages = "15845--15860",
    ISBN = "979-8-89176-332-6"
}

@misc{Bansal2025CRESTUS,
      title={CREST: Universal Safety Guardrails Through Cluster-Guided Cross-Lingual Transfer}, 
      author={Lavish Bansal and Naman Mishra},
      year={2026},
      eprint={2512.02711},
      archivePrefix={arXiv},
      primaryClass={cs.CL},
      url={https://arxiv.org/abs/2512.02711}, 
}

@misc{Verma2025TheHS,
      title={The Hidden Space of Safety: Understanding Preference-Tuned LLMs in Multilingual context}, 
      author={Nikhil Verma and Manasa Bharadwaj},
      year={2025},
      eprint={2504.02708},
      archivePrefix={arXiv},
      primaryClass={cs.CL},
      url={https://arxiv.org/abs/2504.02708}, 
}

@misc{wang2026refusaldirectionuniversalsafetyaligned,
      title={Refusal Direction is Universal Across Safety-Aligned Languages}, 
      author={Xinpeng Wang and Mingyang Wang and Yihong Liu and Hinrich Schütze and Barbara Plank},
      year={2026},
      eprint={2505.17306},
      archivePrefix={arXiv},
      primaryClass={cs.CL},
      url={https://arxiv.org/abs/2505.17306}, 
}

@inproceedings{Shin2026LayerwiseSF,
  title = "Layer-wise Swapping for Generalizable Multilingual Safety",
    author = "Shin, Hyunseo  and
      Hwang, Wonseok",
    editor = "Demberg, Vera  and
      Inui, Kentaro  and
      Marquez, Llu{\'i}s",
    booktitle = "Proceedings of the 19th Conference of the {E}uropean Chapter of the {A}ssociation for {C}omputational {L}inguistics (Volume 1: Long Papers)",
    month = mar,
    year = "2026",
    address = "Rabat, Morocco",
    publisher = "Association for Computational Linguistics",
    url = "https://aclanthology.org/2026.eacl-long.98/",
    doi = "10.18653/v1/2026.eacl-long.98",
    pages = "2223--2238",
    ISBN = "979-8-89176-380-7"
}

@misc{Liang2026MultilingualSA,
      title={Multilingual Safety Alignment Via Sparse Weight Editing}, 
      author={Jiaming Liang and Zhaoxin Wang and Handing Wang},
      year={2026},
      eprint={2602.22554},
      archivePrefix={arXiv},
      primaryClass={cs.LG},
      url={https://arxiv.org/abs/2602.22554}, 
}

@inproceedings{Bell2025TranslateTD,
    title = "Translate, Then Detect: Leveraging Machine Translation for Cross-Lingual Toxicity Classification",
    author = "Bell, Samuel  and
      S{\'a}nchez, Eduardo  and
      Dale, David  and
      Stenetorp, Pontus  and
      Artetxe, Mikel  and
      Costa-Juss{\`a}, Marta R.",
    editor = "Haddow, Barry  and
      Kocmi, Tom  and
      Koehn, Philipp  and
      Monz, Christof",
    booktitle = "Proceedings of the Tenth Conference on Machine Translation",
    month = nov,
    year = "2025",
    address = "Suzhou, China",
    publisher = "Association for Computational Linguistics",
    url = "https://aclanthology.org/2025.wmt-1.15/",
    doi = "10.18653/v1/2025.wmt-1.15",
    pages = "253--268",
    ISBN = "979-8-89176-341-8"
}

@misc{banerjee2026bridgingmultilingualsafetydivide,
      title={Bridging the Multilingual Safety Divide: Efficient, Culturally-Aware Alignment for Global South Languages}, 
      author={Somnath Banerjee and Rima Hazra and Animesh Mukherjee},
      year={2026},
      eprint={2602.13867},
      archivePrefix={arXiv},
      primaryClass={cs.CL},
      url={https://arxiv.org/abs/2602.13867}, 
}

@inproceedings{
Ouyang2022RLHF,
title={Training language models to follow instructions with human feedback},
author={Long Ouyang and Jeffrey Wu and Xu Jiang and Diogo Almeida and Carroll Wainwright and Pamela Mishkin and Chong Zhang and Sandhini Agarwal and Katarina Slama and Alex Gray and John Schulman and Jacob Hilton and Fraser Kelton and Luke Miller and Maddie Simens and Amanda Askell and Peter Welinder and Paul Christiano and Jan Leike and Ryan Lowe},
booktitle={Advances in Neural Information Processing Systems},
editor={Alice H. Oh and Alekh Agarwal and Danielle Belgrave and Kyunghyun Cho},
year={2022},
url={https://openreview.net/forum?id=TG8KACxEON},
address="New Orleans, Louisiana, United States of America"
}

@misc{Bai2022ConstitutionalAI,
      title={Constitutional AI: Harmlessness from AI Feedback}, 
      author={Yuntao Bai and Saurav Kadavath and Sandipan Kundu and Amanda Askell and Jackson Kernion and Andy Jones and Anna Chen and Anna Goldie and Azalia Mirhoseini and Cameron McKinnon and Carol Chen and Catherine Olsson and Christopher Olah and Danny Hernandez and Dawn Drain and Deep Ganguli and Dustin Li and Eli Tran-Johnson and Ethan Perez and Jamie Kerr and Jared Mueller and Jeffrey Ladish and Joshua Landau and Kamal Ndousse and Kamile Lukosuite and Liane Lovitt and Michael Sellitto and Nelson Elhage and Nicholas Schiefer and Noemi Mercado and Nova DasSarma and Robert Lasenby and Robin Larson and Sam Ringer and Scott Johnston and Shauna Kravec and Sheer El Showk and Stanislav Fort and Tamera Lanham and Timothy Telleen-Lawton and Tom Conerly and Tom Henighan and Tristan Hume and Samuel R. Bowman and Zac Hatfield-Dodds and Ben Mann and Dario Amodei and Nicholas Joseph and Sam McCandlish and Tom Brown and Jared Kaplan},
      year={2022},
      eprint={2212.08073},
      archivePrefix={arXiv},
      primaryClass={cs.CL},
      url={https://arxiv.org/abs/2212.08073}, 
}

@misc{zhao2026mlbenchguardpolicygroundedmultilingualsafety,
      title={ML-Bench\&Guard: Policy-Grounded Multilingual Safety Benchmark and Guardrail for Large Language Models}, 
      author={Yunhan Zhao and Zhaorun Chen and Xingjun Ma and Yu-Gang Jiang and Bo Li},
      year={2026},
      eprint={2605.00689},
      archivePrefix={arXiv},
      primaryClass={cs.CL},
      url={https://arxiv.org/abs/2605.00689}, 
}

@misc{zhang2026transferssafetyidentifyingtargeting,
      title={Who Transfers Safety? Identifying and Targeting Cross-Lingual Shared Safety Neurons}, 
      author={Xianhui Zhang and Chengyu Xie and Linxia Zhu and Yonghui Yang and Weixiang Zhao and Zifeng Cheng and Cong Wang and Fei Shen and Tat-Seng Chua},
      year={2026},
      eprint={2602.01283},
      archivePrefix={arXiv},
      primaryClass={cs.CV},
      url={https://arxiv.org/abs/2602.01283}, 
}

@inproceedings{perez-etal-2022-red,
    title = "Red Teaming Language Models with Language Models",
    author = "Perez, Ethan  and
      Huang, Saffron  and
      Song, Francis  and
      Cai, Trevor  and
      Ring, Roman  and
      Aslanides, John  and
      Glaese, Amelia  and
      McAleese, Nat  and
      Irving, Geoffrey",
    editor = "Goldberg, Yoav  and
      Kozareva, Zornitsa  and
      Zhang, Yue",
    booktitle = "Proceedings of the 2022 Conference on Empirical Methods in Natural Language Processing",
    month = dec,
    year = "2022",
    address = "Abu Dhabi, United Arab Emirates",
    publisher = "Association for Computational Linguistics",
    url = "https://aclanthology.org/2022.emnlp-main.225/",
    doi = "10.18653/v1/2022.emnlp-main.225",
    pages = "3419--3448"
}

@inproceedings{yong-etal-2025-state,
    title = "The State of Multilingual {LLM} Safety Research: From Measuring The Language Gap To Mitigating It",
    author = "Yong, Zheng Xin  and
      Ermis, Beyza  and
      Fadaee, Marzieh  and
      Bach, Stephen  and
      Kreutzer, Julia",
    editor = "Christodoulopoulos, Christos  and
      Chakraborty, Tanmoy  and
      Rose, Carolyn  and
      Peng, Violet",
    booktitle = "Proceedings of the 2025 Conference on Empirical Methods in Natural Language Processing",
    month = nov,
    year = "2025",
    address = "Suzhou, China",
    publisher = "Association for Computational Linguistics",
    url = "https://aclanthology.org/2025.emnlp-main.800/",
    doi = "10.18653/v1/2025.emnlp-main.800",
    pages = "15845--15860",
    ISBN = "979-8-89176-332-6"
}

\appendix

\section{Appendix}
\begin{figure*}[h]
    \includegraphics[width=1.0\textwidth]{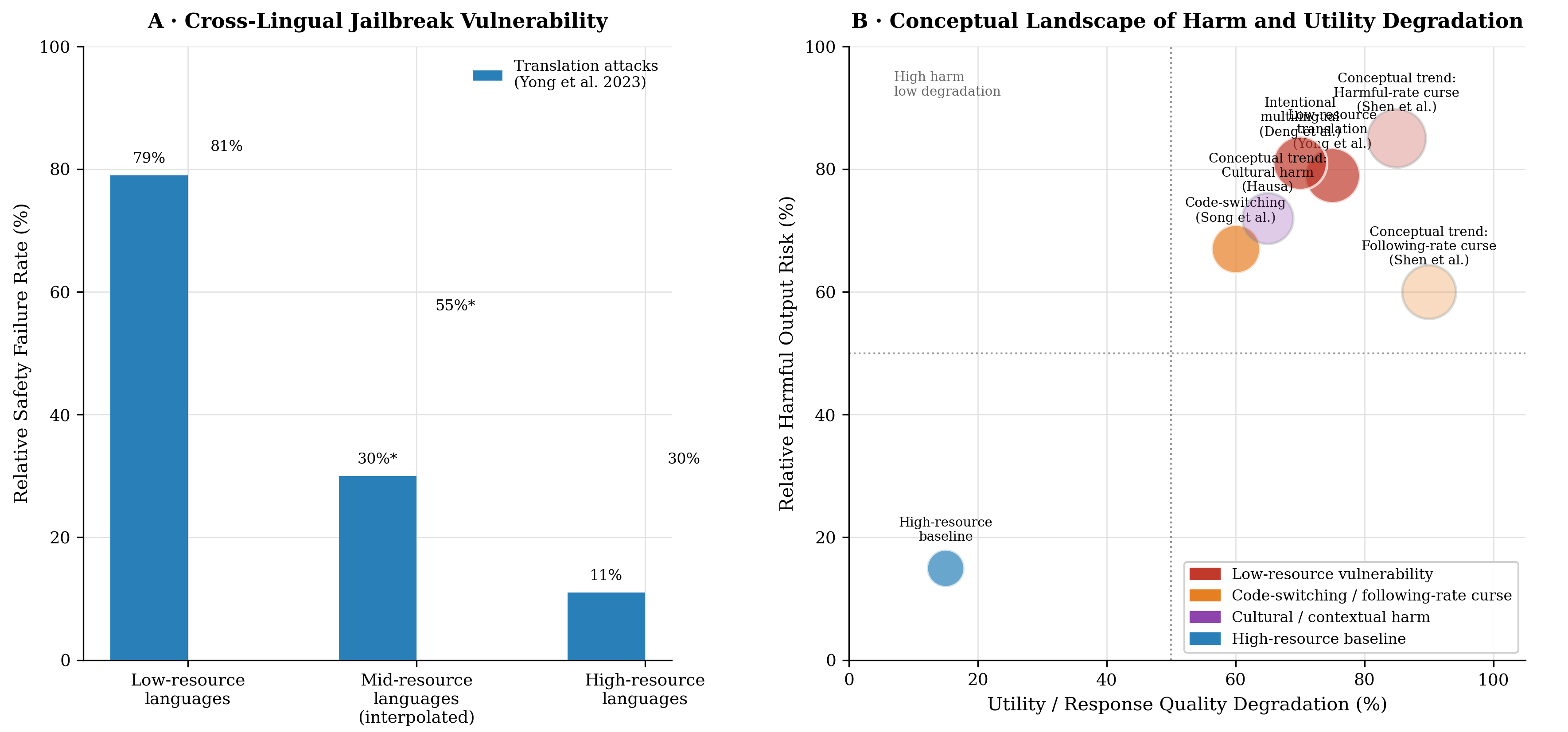}
    \caption{Safety Risks in Multilingual and low-resource settings}
    \label{fig:rq2}
\end{figure*}

\subsection{Search Strings and Criteria}
\label{app:search}

To extract the articles from these different databases, we used the following queries:

\begin{table}[h]
\centering
\small
\caption{Search strings per database.}
\label{tab:search_strings}
\begin{tabular}{p{1.4cm} p{5.8cm}}
\toprule
\textbf{Database} & \textbf{Query (simplified)} \\
\midrule
Semantic Scholar &
  (\textit{LLM} OR \textit{large language model}) AND
  (\textit{safety} OR \textit{jailbreak} OR \textit{toxicity}
  OR \textit{alignment}) AND
  (\textit{multilingual} OR \textit{low-resource} OR
  \textit{African language}) \\
arXiv &
  Same term clusters submitted via arXiv search API;
   \\
OpenAlex &
   Same term clusters submitted via OpenAlex search API\\
\bottomrule
\end{tabular}
\end{table}

\begin{table}[t]
\centering
\small
\caption{Inclusion (I) and exclusion (E) criteria.}
\label{tab:criteria}
\begin{tabular}{p{0.28cm} p{6.8cm}}
\toprule
\multicolumn{2}{l}{\textit{Inclusion}} \\
I1 & Published or preprinted after January 2020. \\
I2 & Written in English. \\
I3 & Primary focus on LLM safety, adversarial robustness,
     toxicity detection, or content moderation. \\
I4 & Covers $\geq$1 non-English, low-resource, or African
     language, or proposes methods applicable to such settings. \\
I5 & Peer-reviewed or substantive preprint with methods and
     empirical results. \\
\midrule
\multicolumn{2}{l}{\textit{Exclusion}} \\
E1 & Safety studied exclusively in English; no multilingual
     component or discussion. \\
E2 & Focused on classical NLP models (LSTM, CNN) with no
     connection to LLMs. \\
E3 & Safety is peripheral, not a primary contribution. \\
E4 & Editorial or opinion piece without empirical results. \\
E5 & Full text inaccessible. \\
\bottomrule
\end{tabular}
\end{table}

\subsection{LLM Usage: LLM Screening Prompt}
\label{app:prompt}

The following prompt was submitted once per candidate record to
Claude Sonnet (claude-sonnet-4.5). No cross-record
context was provided. The JSON output was parsed programmatically;
\texttt{decision} and \texttt{justification} were appended as
columns to the screening spreadsheet.

\begin{quote}
\small\ttfamily
You are a systematic literature review assistant. You will be
given the title, abstract, and metadata of one academic paper.
Assess whether it is relevant to a review on LLM safety alignment
for low-resource and African languages.

Research questions: RQ1 (alignment methods for low-resource
languages), RQ2 (safety risks and cultural harms in multilingual
settings), RQ3 (datasets and benchmarks for low-resource safety),
RQ4 (cross-lingual transfer of safety alignment).

Inclusion: I1 post-2020, I2 English, I3 primary LLM safety
focus, I4 covers or is applicable to low-resource languages,
I5 empirical content.
Exclusion: E1 English-only, E2 non-LLM models, E3 peripheral
safety, E4 no empirical results, E5 inaccessible.

Paper---Title: \{title\}. Year: \{year\}. Venue: \{venue\}.
Abstract: \{abstract\}.

Return ONLY valid JSON:
\{"rqs\_addressed": [...],
 "inclusion\_met": [...],
 "exclusion\_triggered": [...],
 "decision": "keep" | "remove",
 "confidence": "high" | "medium" | "low",
 "justification": "one sentence"\}
\end{quote}

\subsection{Further Analysis}
\begin{figure}[h]
    \includegraphics[width=0.9\textwidth]{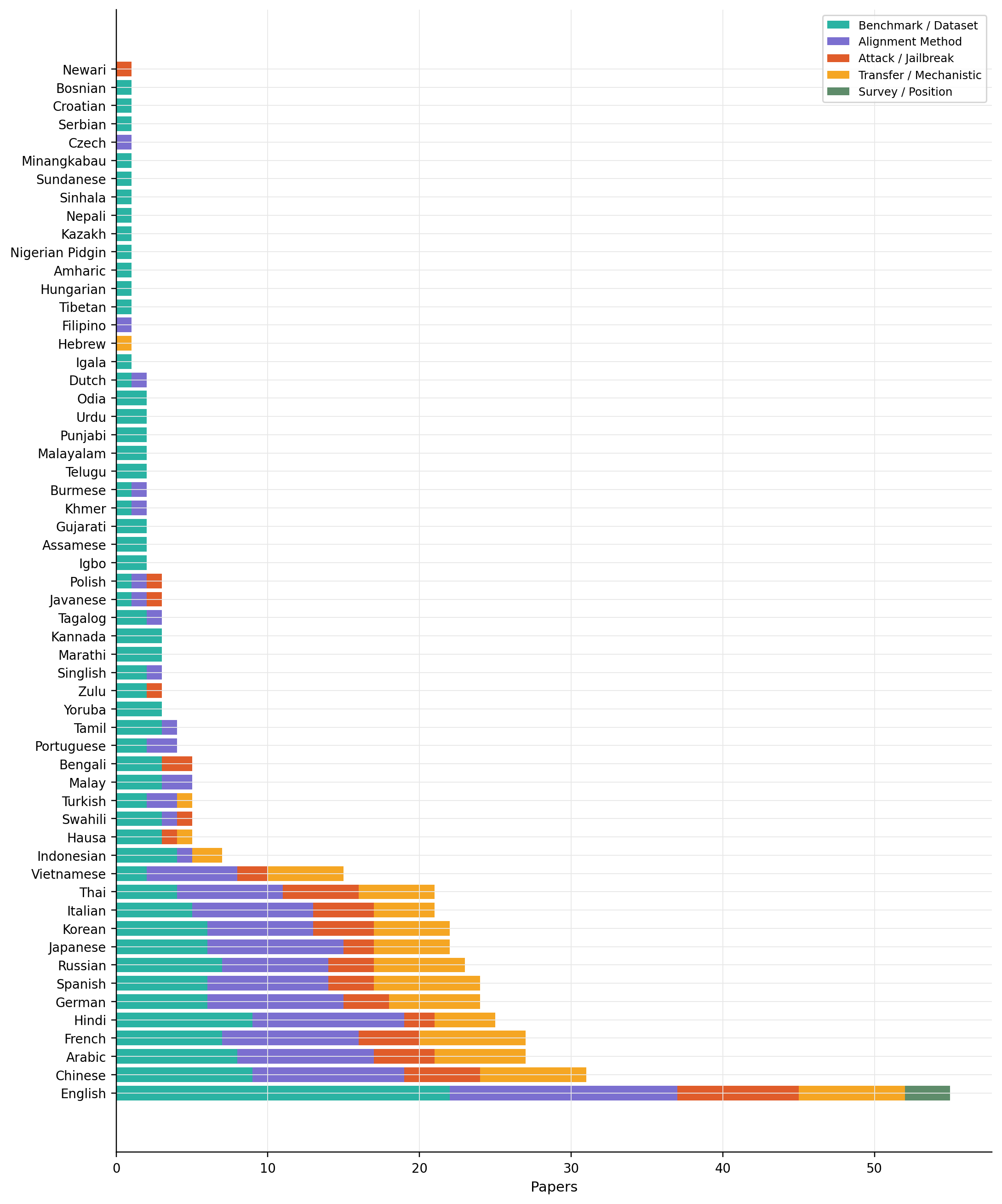}
    \caption{Languages Distribution Across Studies}
    \label{fig:lang}
\end{figure}

The contrast in research coverage illustrated in Figure~\ref{fig:lang} underpins the safety disparities quantified in Figure~\ref{fig:rq2}. As shown in Figure~\ref{fig:lang}, English dominates across all study categories (including benchmarks, alignment methods, and adversarial attacks), while low-resource languages remain confined to a long tail of minimal representation. This imbalance in research attention is reflected in downstream safety performance. Figure~\ref{fig:rq2}(A) shows that low-resource languages exhibit cross-lingual jailbreak failure rates more than seven times higher than those observed in high-resource baselines. Furthermore, Figure~\ref{fig:rq2}(B) highlights a clear separation in the risk–utility landscape: high-resource languages cluster in a region characterized by low harmful output rates and limited utility degradation, whereas low-resource and code-switched settings occupy a high-risk region marked by elevated harmful outputs and substantial utility loss. Together, these results highlight a systematic disparity in safety alignment coverage and underscore the need for more equitable multilingual safety strategies.

\end{document}